\documentclass[runningheads]{llncs}
\usepackage[T1]{fontenc}
\usepackage[export]{adjustbox}
\usepackage{tabularray}
\UseTblrLibrary{booktabs}
\usepackage{amsmath}
\usepackage{xcolor}

\usepackage{graphicx}
\usepackage[pagebackref,breaklinks,colorlinks,allcolors=cvprblue]{hyperref}
\hypersetup{
	breaklinks=true,
	pdfpagemode={UseOutlines}, 
	plainpages=false, 
	pdftitle={Archival Faces: Detection of Faces in Digitized Historical Documents}, 
	pdfauthor={anonymized}, 
	pdfkeywords={}, 
	linkcolor=hrcolor-links, 
	citecolor=hrcolor-cite, 
	urlcolor=hrcolor-urls,
	colorlinks=true
}
\usepackage{color}
\definecolor{hrcolor-links}{HTML}{CC0003}
\definecolor{hrcolor-urls}{HTML}{092EAB}
\definecolor{hrcolor-cite}{HTML}{2F8F00}

\begin{document}
\title{{Archival Faces:} Detection of Faces in Digitized Historical Documents}
\titlerunning{{Archival Faces:} Detection of Faces in Digitized Historical Documents}


\author{Marek Vaško\inst{1}\orcidID{0000-0003-1404-4154} \and
Adam Herout\inst{1}\orcidID{0000-0003-2143-9314} \and
Michal Hradiš\inst{1}\orcidID{0000-0002-6364-129X}}
 \authorrunning{M. Vaško et al.}
\institute{
Faculty of Information Technology, Brno University of Technology, Bozetechova 2/1, 612 00 Brno, Czech Republic \\
\email{\{ivasko,herout,ihradis\}@fit.vutbr.cz}
}

\maketitle              

%
%

\begin{abstract}
When digitizing historical archives, it is necessary to search for the faces of people, especially in newspapers, link them to the surrounding text, and make them searchable. 
Existing face detectors on datasets of scanned historical documents fail remarkably -- current detection tools only achieve around 24\,\% mAP at 50:90\,\% IoU. 
This work compensates for this failure by introducing a new manually annotated domain-specific dataset in the style of the popular Wider Face dataset containing 2.2\,k new images from di\-gi\-ti\-zed historical newspapers from the 19\textsuperscript{th} to 20\textsuperscript{th} century, with 11\,k new bounding-box annotations and associated facial landmarks.  This dataset enables existing detectors to be retrained, bringing their results closer to the standard in the field of face detection in the wild. 
We report several experimental results comparing different families of fine-tuned detectors with publicly available pre-trained face detectors. In ablation studies, we compare multiple detector sizes, providing comprehensive detection and landmark prediction performance results.

\end{abstract}    



\begin{figure}[h]
    \centering
    \includegraphics[height=3.6cm]{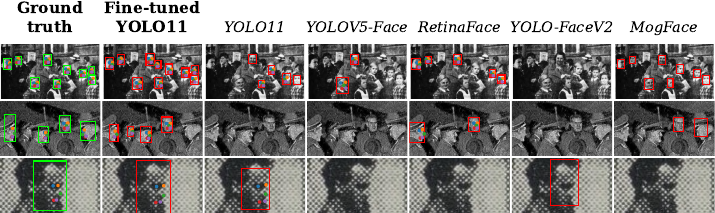}
    \caption{Performance of different state-of-the-art face detectors on historical documents. These chosen images do not appear ambiguous to the human eye; however, detectors fail to predict the proper bounding box.}
    \label{fig:failure}
\end{figure}

\section{Introduction}

Historical archives contain a great deal of interesting information, which, unlike texts produced today, has little digital presence and remains virtually inaccessible to scholars and the public. Advances in writing recognition and detection of non-textual document parts~\cite{monnier_docextractor_2020,pfitzmann_doclaynet_2022,boillet_horae_2019,clausner_icdar2019_2019} make it possible to digitize them and make them accessible in the digital world. Digitization is of interest not only for making archival knowledge accessible to human readers but also for data accessibility in large-scale multimodal language model~\cite{grattafiori_llama_2024,wang_cogvlm_2024} training. However, a sizeable missing link to our knowledge not covered thus far is finding people and their photos in the historical corpus of data. 

Face detection is one `classic' discipline of computer vision \cite{viola_robust_2004,yang_wider_2016,zhang_joint_2016} and an essential stepping stone toward face recognition~\cite{kim_adaface_2022,qin_swinface_2024,dan_transface_2023}. Thanks to the power of deep learning, current face detectors are extremely powerful \cite{vedaldi_end--end_2020,zhao_detrs_2024,bochkovskiy_yolov4_2020,wang_yolov10_2024,tian_yolov12_2025}. Mainly, due to the development of large and variable data sets \cite{yang_wider_2016,zhu_face_2012,phillips_overview_2005,maze_iarpa_2018}, face detection models excel in detecting faces even at very low resolutions~\cite{zhang_robust_2020}, handle occlusions, the presence of distracting objects, a large number of faces in a single image, and other phenomena.

Although the datasets used to train modern face detectors are extensive and varied, they insufficiently represent face images printed by historical printing methods. For this reason, face detectors fundamentally fail on such data, as shown in Figure~\ref{fig:failure}. Even the best pre-trained model, YOLOv5-Face~\cite{qi_yolo5face_2023}, achieves only  $24\,\%$ average precision on facial bounding box prediction in historical documents. We explore this shortcoming of current detectors later in Section~\ref{sec:baselines}. Consequently, these results show a large gap between the recognition of people in historical data and the capabilities of the current state-of-the-art.

Thus, we propose Archival Faces, a face detection dataset from digitized archival materials, to be utilized as a reasonable basis for later research on face recognition in historical documents. The dataset contains annotated bounding boxes of faces and the usual facial landmarks (eyes, mouth corners, nose). Our data set can be used to evaluate face detectors on archival data or for cross-validation experiments. Utilizing cross-validation, we show in Section~\ref{sec:results} how fine-tuning on Archival Faces can significantly improve the detection of faces in historical documents.

The contributions of this paper are as follows:
\begin{enumerate}
    \item We cast light on the detection problem of faces in digitized historical or archival documents and establish a baseline for state-of-the-art detection performance.
    \item We provide a publicly available dataset, Archival Faces\footnote{\href{https://doi.org/10.5281/zenodo.15077975}{doi.org/10.5281/zenodo.15077975}.}, from this underrepresented domain, with landmark annotations to facilitate precise alignment.
    \item We evaluate the performance of generic and Archival Faces fine-tuned face detectors on the Archival Faces.
    \item We provide a detailed ablation with various detector variants, model sizes, fractions of the dataset, and various cross-validation setups. We show that fine-tuning current detectors on as little as $1\,\mathrm{k}$ samples can improve face detection on historical data.
\end{enumerate}

\section{Related work}

\paragraph{Face Detection Datasets.}
The evolution of face detection has been closely tied to the development of comprehensive datasets. Early face datasets such as the XM2VTS database~\cite{messer_xm2vtsdb_1999} and the Face Recognition Grand Challenge~\cite{phillips_overview_2005} were primarily designed for face recognition tasks under controlled conditions, lacking diversity. 
Datasets such as FDDB \cite{jain_fddb_nodate}, PASCAL-face~\cite{yan_face_2014}, COCO Whole-Body~\cite{jin_whole-body_nodate} and Annotated Faces in the Wild~\cite{zhu_face_2012} offer more realistic scenarios, including pose, illumination, and occlusion variations. Commonly used Wider Face dataset~\cite{yang_wider_2016} contains over 32,000 images. 
Despite its scale and diversity, Wider Face and similar datasets predominantly consist of modern digital photographs or video frames. This creates a significant gap in images from earlier periods. Historical images often suffer from issues such as low resolution, grayscale formats, degradation due to aging, and unique artifacts not present in contemporary photos. The performance of models in this domain is not thoroughly explored.

\paragraph{Face Detectors.}
Parallel to the development of datasets, face detection algorithms have undergone significant evolution, particularly with the advent of deep learning. 
Multi-Task Cascaded Convolutional Network (MTCNN)~\cite{zhang_joint_2016} combines face detection with facial landmark localization in a cascade of convolutional neural networks. The S$^3$FD model~\cite{zhang_s3fd_2017} employs a single-shot detection framework with purposely selected anchor scales to effectively capture small faces. PyramidBox~\cite{ferrari_pyramidbox_2018} introduced context-aware mechanisms to incorporate surrounding information. RetinaFace ~\cite{deng_retinaface_2019} extended the single-shot paradigm by incorporating dense regression of facial landmarks and even estimating 3D face structure. Detectors such as the Selective Refinement Network~\cite{chi_selective_2019} and the Dual Shot Face Detector~\cite{li_dsfd_2019} employ multi-step classification and regression schemes to reduce false positives and better localize faces. More recent research has focused on network design through architecture search~\cite{li_asfd_2021,liu_mogface_2022}, adapting existing models to new contexts~\cite{qi_yolo5face_2023,yu_yolo-facev2_2024}, and developing efficient architectures for on-device face detection~\cite{jeong_eresfd_2024,zhu_tinaface_2021,wang_efficientface_2023,guo_sample_2021}

Despite these advances, a critical challenge remains: ensuring that state-of-the-art detectors generalize well to out-of-distribution data, namely historical imagery. 
Although highly accurate on contemporary benchmarks, modern detectors frequently exhibit significant performance drops when applied to archival photographs. 
This is due to the intrinsic differences in image quality, contrast, and artifacts present in digitized historical documents. 
Thus, we show that fine-tuning a detector on as few as $1\,\mathrm{k}$ historical samples can improve performance, surpassing detectors trained on modern datasets. 
This finding validates the need for domain-specific adaptation and emphasizes the importance of integrating specialized datasets like Archival Faces into training pipelines.

\section{Archival Faces: Novel Dataset of Faces from Digitized Historical Documents}

\begin{figure*}[t]
    \centering
    \includegraphics[]{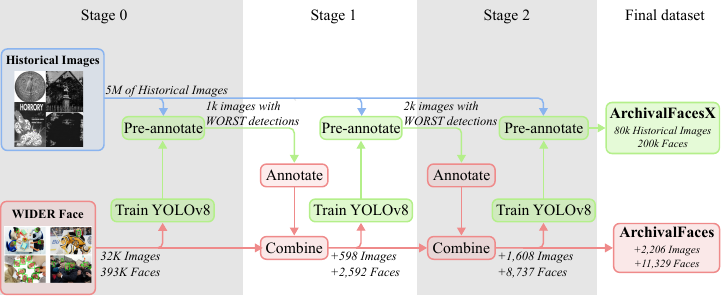}
    \caption{Overview of the annotation pipeline. \textbf{0) Automatic pre-annotation:} one thousand images with the worst detection confidence scores have been selected.  \textbf{1) First human-annotation phase:} bounding boxes on around five hundred images were adjusted or added by annotators; the rest were deemed as not containing any annotations. \textbf{2) Second human-annotation phase:} we have pre-annotated data with a detector trained on data from the previous phase; two thousand images with worst bounding box confidences were then selected.}
    \label{fig:annotation_pipeline}
\end{figure*}

The Archival Faces dataset contains images of people detected in scanned historical newspapers and books with bounding boxes and annotations of facial landmarks. The data set covers multiple printing styles (corresponding to years covered within the archives) and multiple photography styles, such as portraits and crowd photographs. We also have data that contain even statues and caricature drawings, making the data set variable. The source data are completely unlabeled and unprocessed; however, we wanted to utilize as much automated information retrieval from these sources as possible to create the labeled dataset. We mainly leverage existing face-detection networks to retrieve relevant images and human resources for annotation correction.

The annotation of the data set was performed in three distinct phases depicted in Figure~\ref{fig:annotation_pipeline}. The outline of the stages is as follows:

\begin{enumerate}
    \item Historical image dataset extraction. \textbf{Stage~0}
    \item Pre-annotation with a model trained on the Wider Face dataset. \textbf{Stage~0}
    \item First campaign of human annotation. \textbf{Stage~1}
    \item Pre-annotation using human annotated faces from the first human annotation round. \textbf{Stage~1}
    \item Second human annotation round on pre-annotated data from point 4. \textbf{Stage 2}
\end{enumerate}


\subsection{Data Sourcing}

Our work uses a recently digitized corpus from a mixture of Czech libraries and archives from DigitalniKnihovna\footnote{\url{https://www.digitalniknihovna.cz/}}.  Speaking in numbers, the overall size of publicly available scanned document pages is over nine million. Our data sources are overviewed in Figure~\ref{fig:year_hist}, where the majority of images are from \textit{Library Liberec (KVKLI)}\footnote{\url{https://en.kvkli.cz/}}, \textit{Moravian library (MZK)}\footnote{\url{https://www.mzk.cz/}} and \textit{Digital Forum of Middle and East Europe (D)}\footnote{\url{https://www.digitalniknihovna.cz/d}}.

To make annotation more efficient, we considered only image and photograph elements of the sourced document pages. These elements were automatically localized with a detector (YOLOv8~\cite{jocher_ultralytics_2023}) trained on the AnnoPage Dataset~\cite{anonymized_annopage_2025} which contains $7,550$ pages from similar document domain annotated according to official \textit{Methodology of image document processing}~\cite{jirousek_metodika_2024} for Czech libraries.
A face detection confidence-driven sample of the localized visual elements was selected for annotation. 


\subsection{Automatic Face Pre-Extraction and Human Annotation}

\begin{figure}[t]
    \centering
    \includegraphics[]{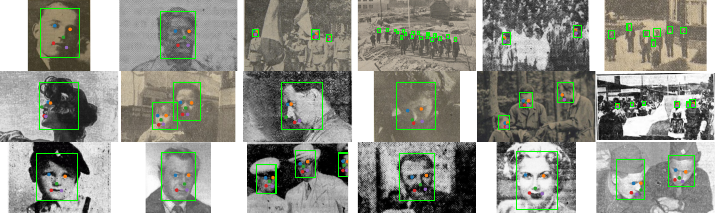}
    \caption{Showcase of facial landmark annotation and annotation of bounding boxes. 
      A random sample of real-world annotations was selected, including different edge cases.}
    \label{fig:facial_landmarks}
\end{figure}

We utilize a pre-trained YOLO v8 model \cite{jocher_ultralytics_2023} trained on the COCO dataset~\cite{lin_microsoft_2014} during our pre-annotation stages (YOLO11 mentioned elsewhere in the paper was not available yet then).  We then train this model on the Wide Face dataset~\cite{yang_wider_2016} utilizing bounding boxes and predictions of facial landmarks. Our dataset follows a similar format proposed by authors of Retina Face~\cite{deng_retinaface_2019}, where we use five landmarks, mainly the middle of the eye sockets, nose tip, and mouth corners, as landmarks during the annotation as seen in Figure~\ref{fig:facial_landmarks}.

Even though the predictions provided by the model trained on Wider Face were relatively poor, we still use them for the $0^\mathrm{th}$ stage dataset pre-annotation to reduce overall annotation time and provide some metrics for annotation candidate selection. 
For the first annotation round (Stage 1 in the figure), we sample data from the entire corpus of historical photos to select images with probability inversely proportional to average face bounding box confidences within the photo $p_{s}(I)=|I|^{-1}\sum_{b \in I}p_{f}(b)$; where $p_{s}(\cdot)$ is the probability of the image sampling, $I$ is the set of bounding boxes detected in the image, $b$ is a bounding box and $p_{f}(\cdot)$ is the probability of bounding box being a face. 
This way, we selected images with faces that were likely challenging to detect and which may benefit the most from human corrections. The human annotator was given the following guidelines for annotation:
\textit{
\begin{enumerate}
    \item Remove any false positives and duplicate annotations.
    \item Annotate landmarks for eyes, nose tip, and mouth corners with appropriate landmark classes. Fix any miss-aligned and add any missing annotations. If a person is facing away from the camera, try to estimate the position of landmarks.
    \item Annotate bounding boxes that contain the entire face and are tightly around the face outline defined by ears, chin, and hairline. 
\end{enumerate}
}
The pre-annotation model was then re-trained on $80\,\%$ of human-annotated historical data ($20\,\%$ was kept for validation) and the training portion of the Wider Face dataset (end of Stage 1). New data for labeling was then selected with the same weighting as in the $0^\mathrm{th}$ round, and we sampled $2k$ new images. The annotation round was then repeated (Stage 2). The overall annotation and pre-annotation pipeline is shown in Figure~\ref{fig:annotation_pipeline}.

\subsection{Overview of Dataset Properties}

\begin{figure}[p]
    \centering
    \includegraphics[]{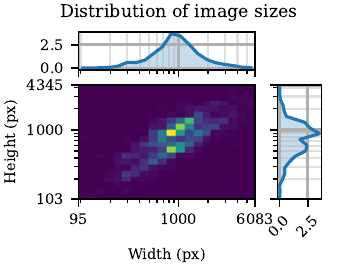}
    \includegraphics[]{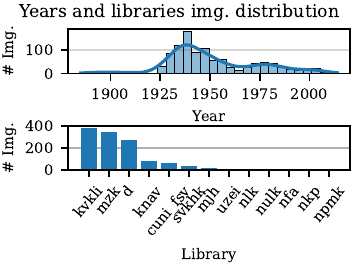}
    \caption{\textbf{Left:} 2-dimensional histogram of image sizes in logarithmic scale. \textbf{Top-right:} Histogram source document publishing years (where available). \textbf{Bottom-right:} \textit{Czech libraries} sources within the dataset; only library abbreviations are reported.}
    \label{fig:year_hist}
\end{figure}

\begin{figure}[p]
    \centering
    \includegraphics[]{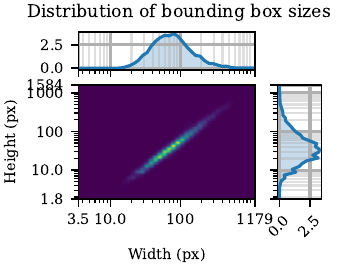}
    \includegraphics[]{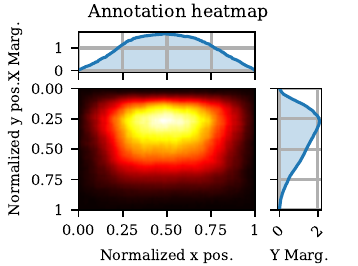}
    \caption{\textbf{Left:} 2D histogram of bounding box sizes in pixels. \textbf{Right:} normalized 2D histogram of annotated image portions, taken as a sum over all normalized bounding-boxes positions.}
    \label{fig:bboc_hist}
\end{figure}

\begin{figure}[p]
    \centering
    \includegraphics[]{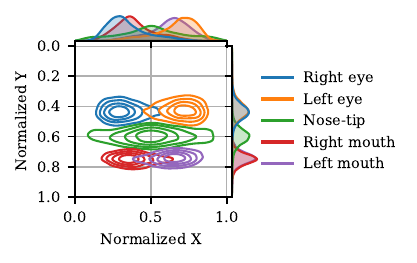}
    \includegraphics[]{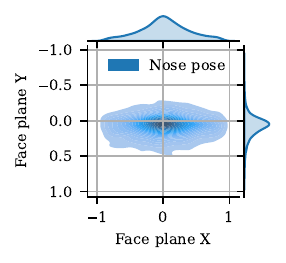}
    \caption{\textbf{Left:} histogram of normalized keypoint positions within bounding box coordinates. \textbf{Right:} distribution of nose position within coordinates of face plane, showing approximate head pose distribution.}
    \label{fig:poses}
\end{figure}

The resulting dataset consists of over $2,206$ annotated historical images containing faces. Since the annotation process began with three thousand images, the dataset has $794$ pictures confirmed to have no face (hard background images). Statistics of the dataset in Figure~\ref{fig:year_hist} report basic information about the input data, such as the distribution of image sizes, distribution of images by years, and the distribution of source libraries represented in the dataset. Interestingly, most image sizes are within the $1000\mathrm{px}$ range, and aspect ratios are mostly landscape. The distribution of document publishing years showcases that most images are over 60 to 100 years old. The dataset contains a tail end of data towards the contemporary era. 

In Figure~\ref{fig:bboc_hist}, we showcase both distributions of bounding box sizes, where they can be seen as Gaussian on a logarithmic width-height scale. Sizes of most of the face bounding boxes are around 50 to 100 px; we also have data represented in tiny bounding box sizes of at least 10 px by each size. The right side of this figure showcases how the different annotations are located in normalized image coordinates. This shows that most facial bounding boxes are widely spread around the center of the image and shifted more toward the top of the picture, which makes sense in the context of full-body and neck-up portraits. 

Figure~\ref{fig:poses} (left) illustrates the distribution of facial key points within the face bounding box, with most of them positioned as if the subject were looking directly at the camera. However, pose variations exist, as shown on the right side, where we plot the nose tip position within the face plane.

To define the face plane, we use two vectors: (1) the average of the eye-to-eye and mouth corner-to-corner lines, and (2) the average of the vectors from each mouth corner to its corresponding eye. These vectors form a matrix, which we invert and use to transform the nose-tip position. This provides a rough pose estimation, assuming the nose extends outside the face plane in 3D. Poses cluster around the zero point, with significant side-to-side variation but minimal vertical deviations.

\section{Dataset Evaluation Protocol}
\label{sec:eval}

The overall approach to evaluation is as follows: Concerning a relatively small amount of annotated data in the dataset, the primary way of the assessment is on the entire dataset. This can be done using a model trained on different facial datasets or by training a model using cross-validation on any defined cross-validation setup. For this purpose, the Archival Faces dataset is split into 10 parts labeled $i=0\dots9$; the data within the splits was selected by randomly reordering the sorted list of annotated images $I_{\mathrm{rand}}=\mathrm{shuffle}(I_{\mathrm{sorted}})$. Then, for each $i^\mathrm{th}$ fold, we select every $10^\mathrm{th}$ element starting from the $i^\mathrm{th}$ element of the shuffled list. Statistics from each fold concerning the number of annotations, distribution across the years, and number of images are reported in Table~\ref{tab:folds_stats}. The number of pictures in each split is $220.6\pm0.6$, with the majority of the difference arising from 2 different applications for splitting after each annotation round. The number of annotations within each split is $1.13k\pm0.1k$, where the difference mainly depends on the number of annotations per specific image selected in the folds.

\begin{table}[t]
    \centering
    \caption{Detailed statistics of data contained within each cross-validation split of the Archival Faces dataset. We report the distribution of images across the years, the count of images, and the count of annotated bounding boxes within each split.}
    \begin{tblr}{
        width=0.49\textwidth,
        colspec = {X[c, m] | X[c,m] | X[c,m] | X[1.8cm,c,h]},
        stretch = 0.0,
        rowsep = 1.1pt,
        hline{1,Z} = {1.2pt},
        hline{2} = {0.5pt},
        hline{3-Y} = {0.5pt,dashed},
    }
        \textbf{Split} & \textbf{Img.} & \textbf{Ann.} & \textbf{Year dis.} \\
        \textbf{0} & 221 & 1074 & \includegraphics[]{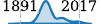}  \\
        \textbf{1} & 220 & 1357 & \includegraphics[]{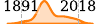}  \\
        \textbf{2} & 221 & 1247 & \includegraphics[]{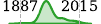}  \\
        \textbf{3} & 221 & 1090 & \includegraphics[]{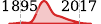}  \\
        \textbf{4} & 221 & 1130 & \includegraphics[]{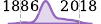}  \\
    \end{tblr}
    \begin{tblr}{
        width=0.49\textwidth,
        colspec = {X[c, m] | X[c,m] | X[c,m] | X[1.8cm,c,h]},
        stretch = 0.0,
        rowsep = 1.1pt,
        hline{1,Z} = {1.2pt},
        hline{2} = {0.5pt},
        hline{3-Y} = {0.5pt,dashed},
    }
        \textbf{Split} & \textbf{Img.} & \textbf{Ann.} & \textbf{Year dis.} \\
        \textbf{5} & 221 &  1052 & \includegraphics[]{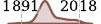}  \\
        \textbf{6} & 221 &  978 & \includegraphics[]{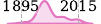}  \\
        \textbf{7} & 221 &  1200 & \includegraphics[]{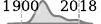}  \\
        \textbf{8} & 220 &  1186 & \includegraphics[]{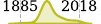}  \\
        \textbf{9} & 219 &  1015 & \includegraphics[]{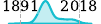}  \\
    \end{tblr}
    \label{tab:folds_stats}
\end{table}

One may either evaluate on all splits together and use them as a test-only dataset or perform $k=10$, $k=5$, or $k=2$ cross-validation. 
The splits included in the cross-validation test sets are always from $i$ to $i+k-1$, where $i=j\times k$, for each $j$-th cross-validation experiment. 
The rest of the splits are designated for the development of the model and can be split into training or validation in any fashion. 
The results are computed according to the standard evaluation protocol defined in the \textit{COCO} Object detection challenge~\cite{lin_microsoft_2014}. We are mainly interested in two types of metrics, one for the bounding box prediction and the other for facial landmark prediction. 
The tested detectors can predict bounding boxes, landmarks, or both, but only relevant statistics for the detected modality should be reported.

\paragraph{Bounding-box detection metrics.} The minimum reporting requirement is average precision (AP) for face bounding boxes.  Intersection over Union ($\mathrm{IoU}$) thresholds are fixed at $0.5$ and $0.75$. The main metric used for model-to-model comparison is an average of $\mathrm{mAP}$ from $\mathrm{IoU}$ sweep between $0.5$ to $0.95$ with step size $0.05$.

\paragraph{Facial landmarks prediction metrics.} Almost all bounding boxes in the dataset have five corresponding landmark predictions. Most are labeled as ``possible to predict''. For each landmark where the visibility is set as \textbf{visible}, the metrics reported are equivalent to \textit{COCO} keypoint prediction task metrics. Similarly to bounding boxes, the mean-average precision should be reported. $\mathrm{IoU}$ used with bounding boxes is replaced by Object Keypoint Similarity ($\mathrm{OKS}$), and equivalent $\mathrm{OKS} \approx \mathrm{IoU}$ thresholds are reported. In this dataset, we do not have any redundantly annotated images; thus, we follow the current version of Ultralitics YOLO~\cite{jocher_ultralytics_2024} package where all values of per-keypoint $\mathrm{OKS}$ parameter $k_i$ are set to $0.2$.

\section{Baselines -- Performance of the Off-the-Shelf Detectors}

\label{sec:baselines}
\begin{table}[p]
\centering
\caption{Detection performance of the state-of-the-art models on the whole Archival Faces dataset, ordered from best to worst. We report mAP at IoU or OKS thresholds defined within the evaluation protocol. The best overall performing model is \textit{YOLO5Face}~\cite{qi_yolo5face_2023}, although none reach above $0.25$ mean-average precision for bounding boxes and $0.47$ for key points. We also include performance for \textit{YOLOv8} and \textit{YOLO11}, which we fine-tuned on Wider Face~\cite{yang_wider_2016}.}
\label{tab:sota_bbox}
\scriptsize
\begin{tblr}{
    width=0.49\textwidth,
    colspec = {X[r, b] | Q[c,m] | Q[c,m] | Q[c,m] | Q[c,m] | Q[c,m] | Q[c,m] | Q[c,m] },
    stretch = 0.3,
    rowsep = 1.1pt,
    vline{2,3} = {1,31}{text=\ },
    vline{4,5} = {3}{text=\ },
    vline{7,8} = {5}{text=\ },
    hline{2,32,Z} = {1.2pt},
    hline{4,6} = {0.5pt},
    hline{7-31,33-Y} = {0.5pt,dashed},
    cell{1}{3} = {c = 6, r = 1}{c,m},
    cell{2}{1} = {c = 1, r = 4}{r,f},
    cell{2}{2} = {c = 1, r = 4}{r,f},
    cell{2}{3} = {c = 6, r = 1}{c,m},
    cell{3}{6} = {c = 3, r = 1}{c,m},
    cell{4}{3} = {c = 6, r = 1}{c,b},
    cell{5}{3} = {c = 3, r = 1}{c,b},
    cell{31}{1} = {c = 2, r = 1}{c,m},
    cell{31}{3} = {c = 6, r = 1}{c,m},
}

&  & \textbf{Bounding box mAP} \\
\textbf{Model} & \textbf{Variant} & \textbf{IoU/OKS} \\
 & & {50:95} & {75} & {50} & {50:95} \\
 & & \textbf{Scales} \\
 & & {all} & & & {s.} & {m.} & {l.} \\

\textit{YOLOV5-Face~\cite{qi_yolo5face_2023}} & \textit{M} & 0.24 & 0.19 & 0.52 & 0.20 & 0.22 & 0.33 \\
\textit{YOLOV5-Face~\cite{qi_yolo5face_2023}} & \textit{S} & 0.22 & 0.17 & 0.48 & 0.19 & 0.20 & 0.29 \\
\textit{YOLO11} & \textit{L} & 0.22 & 0.13 & 0.53 & 0.18 & 0.20 & 0.29 \\
\textit{YOLOV5-Face~\cite{qi_yolo5face_2023}} & \textit{N} & 0.22 & 0.18 & 0.46 & 0.18 & 0.20 & 0.30 \\
\textit{YOLO11} & \textit{M} & 0.20 & 0.11 & 0.49 & 0.17 & 0.18 & 0.27 \\
\textit{YOLOv8} & \textit{L} & 0.20 & 0.11 & 0.50 & 0.16 & 0.17 & 0.28 \\
\textit{RetinaFace~\cite{deng_retinaface_2019}} & \textit{-} & 0.20 & 0.12 & 0.46 & 0.15 & 0.17 & 0.29 \\
\textit{YOLOv8} & \textit{M} & 0.19 & 0.10 & 0.48 & 0.15 & 0.17 & 0.26 \\
\textit{YOLOV5-Face~\cite{qi_yolo5face_2023}} & \textit{N 0} & 0.18 & 0.14 & 0.40 & 0.15 & 0.18 & 0.23 \\
\textit{YOLOv8} & \textit{S} & 0.18 & 0.09 & 0.45 & 0.15 & 0.16 & 0.25 \\
\textit{YOLOv8} & \textit{N} & 0.17 & 0.08 & 0.45 & 0.13 & 0.16 & 0.23 \\
\textit{YOLO11} & \textit{N} & 0.17 & 0.08 & 0.43 & 0.12 & 0.16 & 0.22 \\
\textit{YOLO11} & \textit{S} & 0.17 & 0.08 & 0.43 & 0.15 & 0.15 & 0.23 \\
\textit{MogFace~\cite{liu_mogface_2022}} & \textit{SSE} & 0.17 & 0.06 & 0.47 & 0.06 & 0.16 & 0.28 \\
\textit{ASFD~\cite{li_asfd_2021}} & \textit{-} & 0.16 & 0.06 & 0.48 & 0.09 & 0.16 & 0.26 \\
\textit{YOLO-FaceV2~\cite{yu_yolo-facev2_2024}} & \textit{-} & 0.16 & 0.07 & 0.44 & 0.13 & 0.14 & 0.24 \\
\textit{MogFace~\cite{liu_mogface_2022}} & \textit{Ali-AMS} & 0.16 & 0.05 & 0.48 & 0.07 & 0.16 & 0.24 \\
\textit{MogFace~\cite{liu_mogface_2022}} & \textit{E} & 0.15 & 0.08 & 0.39 & 0.02 & 0.14 & 0.29 \\
\textit{MogFace~\cite{liu_mogface_2022}} & \textit{-} & 0.14 & 0.05 & 0.43 & 0.08 & 0.14 & 0.23 \\
\textit{EResFD~\cite{jeong_eresfd_2024}} & \textit{-} & 0.12 & 0.06 & 0.31 & 0.14 & 0.11 & 0.12 \\
\textit{YuNet~\cite{wu_yunet_2023}} & \textit{int8bq} & 0.12 & 0.06 & 0.31 & 0.12 & 0.11 & 0.13 \\
\textit{YuNet~\cite{wu_yunet_2023}} & \textit{-} & 0.11 & 0.06 & 0.30 & 0.12 & 0.11 & 0.13 \\
\textit{YuNet~\cite{wu_yunet_2023}} & \textit{int8} & 0.11 & 0.06 & 0.30 & 0.11 & 0.11 & 0.13 \\
\textit{LightDSFD~\cite{jian_lightdsfd_2024}} & \textit{-} & 0.07 & 0.03 & 0.20 & 0.02 & 0.07 & 0.11 \\
\textit{MTCNN~\cite{zhang_joint_2016}} & \textit{-} & 0.05 & 0.02 & 0.14 & 0.04 & 0.05 & 0.08 \\

&  & {\ \\ \textbf{Keypoint mAP}} \\
\textit{YOLOV5-Face~\cite{qi_yolo5face_2023}} & \textit{M} & 0.47 & 0.47 & 0.48 & 0.47 & 0.67 & 0.50 \\
\textit{RetinaFace~\cite{deng_retinaface_2019}} & \textit{-} & 0.44 & 0.44 & 0.44 & 0.41 & 0.64 & 0.44 \\
\textit{YOLO11} & \textit{L} & 0.44 & 0.44 & 0.44 & 0.45 & 0.63 & 0.45 \\
\textit{YOLOV5-Face~\cite{qi_yolo5face_2023}} & \textit{S} & 0.43 & 0.43 & 0.44 & 0.44 & 0.58 & 0.46 \\
\textit{YOLOv8} & \textit{L} & 0.41 & 0.41 & 0.41 & 0.42 & 0.62 & 0.42 \\
\textit{YOLOV5-Face~\cite{qi_yolo5face_2023}} & \textit{N} & 0.40 & 0.41 & 0.42 & 0.40 & 0.58 & 0.43 \\
\textit{YOLOv8} & \textit{N} & 0.40 & 0.40 & 0.41 & 0.43 & 0.55 & 0.41 \\
\textit{YOLOv8} & \textit{M} & 0.39 & 0.39 & 0.39 & 0.39 & 0.60 & 0.40 \\
\textit{YOLO11} & \textit{M} & 0.38 & 0.39 & 0.39 & 0.38 & 0.58 & 0.40 \\
\textit{YOLOv8} & \textit{S} & 0.38 & 0.38 & 0.39 & 0.39 & 0.57 & 0.39 \\
\textit{YOLO11} & \textit{N} & 0.37 & 0.37 & 0.38 & 0.39 & 0.54 & 0.38 \\
\textit{YOLOV5-Face~\cite{qi_yolo5face_2023}} & \textit{N 0} & 0.35 & 0.36 & 0.37 & 0.37 & 0.47 & 0.38 \\
\textit{YOLO11} & \textit{S} & 0.35 & 0.35 & 0.35 & 0.35 & 0.54 & 0.36 \\
\textit{YuNet~\cite{wu_yunet_2023}} & \textit{int8bq} & 0.27 & 0.27 & 0.28 & 0.27 & 0.30 & 0.27 \\
\textit{YuNet~\cite{wu_yunet_2023}} & \textit{int8} & 0.27 & 0.27 & 0.27 & 0.27 & 0.29 & 0.27 \\
\textit{YuNet~\cite{wu_yunet_2023}} & \textit{-} & 0.26 & 0.27 & 0.27 & 0.27 & 0.30 & 0.27 \\
\textit{MTCNN~\cite{zhang_joint_2016}} & \textit{-} & 0.09 & 0.11 & 0.15 & 0.09 & 0.15 & 0.10 \\
\end{tblr}
\end{table}

We test multiple publicly available face detectors to understand how well current state-of-the-art detectors generalize to the Archival Faces dataset. We compare the results of these detectors and evaluate the mean-average precision according to the evaluation protocol from Section~\ref{sec:eval} for both bounding boxes and key points where applicable. In the comparison, we include mainly detectors fine-tuned on the Wider~Face~\cite{yang_wider_2016} dataset. 

The selection of detectors for comparison was based on the availability of their source code and of the pre-trained weights; we took into account on how recent the approaches are. 
The comparison includes different versions of existing detectors transfer-learned to the face detection task~\cite{qi_yolo5face_2023,yu_yolo-facev2_2024}, task-specific face detectors with custom architectures~\cite{liu_mogface_2022,deng_retinaface_2019,zhang_joint_2016} and neural architecture searched face detector architectures~\cite{li_asfd_2021}. 
We also compare results from low-latency models, which were fine-tuned to specifically work on face detection~\cite{wu_yunet_2023,jian_lightdsfd_2024,jeong_eresfd_2024}. 

In the latter ablation, we use fine-tuning of generic detector architectures. This com\-pa\-ri\-son thus includes baseline results for all sizes of YOLO11~\cite{jocher_ultralytics_2024} and YO\-LO\-v8~\cite{jocher_ultralytics_2023} models. In both cases, we utilize the training setup mentioned in Section~\ref{sec:results}. The only difference is that we use just the Wider Face dataset for model fine-tuning but keep the rest of the recipe the same.

The precisions of bounding-box and keypoint predictions are reported in Table~\ref{tab:sota_bbox}. The table's key point and bounding box precision are ordered from best to worst. We mainly interpret the results for mean-average precision on $\mathrm{IoU}$ and $\mathrm{OKS}$ of $0.5:0.95$. The presented results
 show that the best model YOLOV5-Face~\cite{qi_yolo5face_2023} can achieve $24\%$ bounding box $\mathrm{mAP}$, and $47\%$ keypoint $\mathrm{mAP}$. 
We show that this can be easily surpassed in the next section of the paper. Interestingly, recent off-the-shelf YOLO detectors, which we ``just'' fine-tuned on Wider Face, placed in around third place with YOLO11, where the gap from the best is only around $2\,\%$.

\section{Experiments}
\label{sec:results}

Our experimental evaluation aims to show that fine-tuning recent state-of-the-art off-the-shelf detectors can improve face detection on the Archival Faces dataset. We mostly demonstrate how different model sizes and versions of YOLO detectors~\cite{jocher_ultralytics_2023,jocher_ultralytics_2024} perform when evaluated using cross-validation on the Archival Faces dataset. All experiments mentioned here follow the training setup discussed below. In the latter subsection, we mainly summarize different ablations focused on model and data configurations. In addition to standard YOLO detectors in the last section, we showcase the performance of fine-tuned detection transformers~\cite{vedaldi_end--end_2020,meng_conditional_2021}.

\paragraph{Data.} During the training, we utilize Wider Face datasets and any selected portions of Archival Faces. From the preliminary experiments, we found out that appending Wider Face to domain-specific Archival Faces can improve the overall performance by up to $2\,\%\ \mathrm{mAP}$ compared to the Archival Faces only trained model. Thus, for all experiments, we train our model on around $13\,\mathrm{k}$ images from Wider Face~\cite{yang_wider_2016} and $2\,\mathrm{k}$ (number depends on selected cross-validation setup) from Archival Faces. All major experiments are done on 10-fold cross-validation as defined in Section~\ref{sec:eval}. In other ablations, we try different cross-validation setups for the number of folds. Validation during training is done on one randomly selected section of the designated train portion of the Archival Faces dataset. In all experiments with YOLO detectors, the images are scaled to $640\times640$; we utilize standard augmentation such as random scale, mosaic~\cite{bochkovskiy_yolov4_2020}, and random color jitter.

\paragraph{Model training setup.} In most of our experiments, we fine-tune COCO~\cite{lin_microsoft_2014} pre-trained YOLO detector to the experiment-specific training dataset. All models use the training recipe provided in Ultralytics package~\cite{ultralytics_yolov5_nodate,jocher_ultralytics_2023,jocher_ultralytics_2024}. Device batch size depends on the model size (for nano $32$, small $24$, medium $20$, and large $16$); we utilize automatic mixed-precision training. We use the generic augmentations mentioned above, and the learning rate is set automatically with a scale-up scale-down scheduler. We set the maximum number of epochs to $120$; however, in most cases, we have not seen any improvement on the validation portion of the dataset after $\approx60$ epochs. The model used in the evaluation is always selected according to the best overall validation loss value.

\begin{table}[t]
\centering
\caption{Mean-average precision for bounding box and keypoint predictions after fine-tuning current state-of-the-art detection models (YOLOv8, YOLO11) on $k=10$ cross-validation Archival Faces setup. Each training dataset includes $\approx90\,\%$ of the Archival faces dataset; the rest is used for testing. Results are ordered best-to-worst according to $\mathrm{mAP}$ for $\mathrm{IoU}$ and $\mathrm{OKS}$ sweep between $0.5:0.95$.}
\scriptsize
\begin{tblr}{
    width=0.49\textwidth,
    colspec = {X[r, b] | Q[c,m] | Q[c,m] | Q[c,m] | Q[c,m] | Q[c,m] | Q[c,m] | Q[c,m] },
    stretch = 0.3,
    rowsep = 1.1pt,
    vline{2,3} = {1}{text=\ },
    vline{2,3} = {15}{text=\ },
    vline{4,5} = {3}{text=\ },
    vline{7,8} = {5}{text=\ },
    hline{2,16,Z} = {1.2pt},
    hline{4,6} = {0.5pt},
    hline{7-15,17-Y} = {0.5pt,dashed},
    cell{1}{3} = {c = 6, r = 1}{c,m},
    cell{2}{1} = {c = 1, r = 4}{r,f},
    cell{2}{2} = {c = 1, r = 4}{r,f},
    cell{2}{3} = {c = 6, r = 1}{c,m},
    cell{3}{6} = {c = 3, r = 1}{c,m},
    cell{4}{3} = {c = 6, r = 1}{c,b},
    cell{5}{3} = {c = 3, r = 1}{c,b},
    cell{15}{3} = {c = 6, r = 1}{c,f},
}

&  & \textbf{Bounding box mAP} \\
\textbf{Model} & \textbf{Variant} & \textbf{IoU or OKS} \\
 & & {50:95} & {75} & {50} & {50:95} \\
 & & \textbf{Scales} \\
 & & {all} & & & {s.} & {m.} & {l.} \\

\textit{YOLO11} & \textit{L} & 0.54 & 0.60 & 0.82 & 0.40 & 0.56 & 0.62 \\
\textit{YOLO11} & \textit{M} & 0.53 & 0.60 & 0.82 & 0.40 & 0.56 & 0.62 \\
\textit{YOLOv8} & \textit{L} & 0.53 & 0.58 & 0.81 & 0.39 & 0.55 & 0.62 \\
\textit{YOLOv8} & \textit{M} & 0.52 & 0.57 & 0.80 & 0.38 & 0.54 & 0.60 \\
\textit{YOLO11} & \textit{S} & 0.51 & 0.56 & 0.79 & 0.37 & 0.53 & 0.61 \\
\textit{YOLOv8} & \textit{S} & 0.50 & 0.54 & 0.78 & 0.35 & 0.52 & 0.59 \\
\textit{YOLOv8} & \textit{N} & 0.46 & 0.50 & 0.75 & 0.31 & 0.49 & 0.57 \\
\textit{YOLO11} & \textit{N} & 0.45 & 0.48 & 0.73 & 0.30 & 0.47 & 0.56 \\
\textit{YOLOV5-Face~\cite{qi_yolo5face_2023}} & \textit{M} & \textit{0.24} & \textit{0.19} & \textit{0.52} & \textit{0.20} & \textit{0.22} & \textit{0.33} \\

& & {\ \\\textbf{Keypoints mAP}} \\

\textit{YOLO11} & \textit{L} & 0.72 & 0.73 & 0.73 & 0.76 & 0.93 & 0.74 \\
\textit{YOLO11} & \textit{M} & 0.72 & 0.73 & 0.73 & 0.76 & 0.93 & 0.74 \\
\textit{YOLOv8} & \textit{L} & 0.71 & 0.72 & 0.72 & 0.75 & 0.92 & 0.73 \\
\textit{YOLOv8} & \textit{M} & 0.71 & 0.71 & 0.72 & 0.74 & 0.91 & 0.72 \\
\textit{YOLO11} & \textit{S} & 0.70 & 0.71 & 0.71 & 0.74 & 0.91 & 0.72 \\
\textit{YOLOv8} & \textit{S} & 0.69 & 0.70 & 0.70 & 0.73 & 0.90 & 0.71 \\
\textit{YOLOv8} & \textit{N} & 0.67 & 0.67 & 0.68 & 0.70 & 0.89 & 0.68 \\
\textit{YOLO11} & \textit{N} & 0.65 & 0.66 & 0.67 & 0.69 & 0.85 & 0.67 \\
\textit{YOLOV5-Face~\cite{qi_yolo5face_2023}} & \textit{M} & \textit{0.47} & \textit{0.47} & \textit{0.48} & \textit{0.47} & \textit{0.67} & \textit{0.50} \\

\end{tblr}

\label{tab:finetuning_k_10}
\end{table}

\subsection{State-of-the art detection model fine-tuning}

As part of the experimentation, we have tried two different model variants, YOLOv8~\cite{jocher_ultralytics_2023} and YOLO11~\cite{jocher_ultralytics_2024}. For both architectures, we utilize the model sizes nano (N), small (S), medium (M), and large (L). Results for each model are presented in Table~\ref{tab:finetuning_k_10}. The main conclusion of these experiments is that overall, YOLO11 models perform better for both keypoint and bounding box prediction tasks when compared to YOLOv8, the only exception being the nano model variants where YOLOv8 has $1\%$ to $4\%$ advantage depending on the metric. Also, the higher parameter count model variants perform better. In contrast, the nano variant performs consistently the worst, and the large and medium variants are more on par with each other. Fine-tuned models provide as much as $125\,\%$ increase in $\mathrm{mAP}$ (for $\mathrm{IoU}$ $50:95$) compared to the best generic YOLOV5-Face~\cite{qi_yolo5face_2023} face detector.

\subsection{Ablation of different cross-validation setups}
\begin{table}[t]
\centering
\caption{Results for bounding box prediction performance of YOLO11~\cite{jocher_ultralytics_2024} Large fine-tuned on different cross-validation configurations, where $k=2$, $k=5$ and $k=10$ present how many folds of the datasets were used for each experiment. We ablate the effect of the training model with a smaller amount of data within the training set. }
\scriptsize
\begin{tblr}{
    width=0.49\textwidth,
    colspec = {Q[c,f,1.7cm] | Q[c,f,1.5cm] | Q[c,m] | Q[c,m] | Q[c,m] | Q[c,m] | Q[c,m] | Q[c,m] },
    stretch = 0.3,
    rowsep = 1.1pt,
    vline{2,3} = {1}{text=\ },
    vline{2,3} = {9}{text=\ },
    vline{4,5} = {3}{text=\ },
    vline{7,8} = {5}{text=\ },
    hline{2,10,Z} = {1.2pt},
    hline{4,6} = {0.5pt},
    hline{7-9,11-Y} = {0.5pt,dashed},
    cell{1}{1} = {c = 2, r = 1}{c,m},
    cell{1}{3} = {c = 6, r = 1}{c,m},
    cell{2}{1} = {c = 1, r = 4}{r,m},
    cell{2}{2} = {c = 1, r = 4}{r,m},
    cell{2}{3} = {c = 6, r = 1}{c,m},
    cell{3}{6} = {c = 3, r = 1}{c,m},
    cell{4}{3} = {c = 6, r = 1}{c,b},
    cell{5}{3} = {c = 3, r = 1}{c,b},
    cell{9}{1} = {c = 2, r = 1}{c,f},
    cell{9}{3} = {c = 6, r = 1}{c,f},
}

\textbf{YOLO11 Large} &  & \textbf{Bounding box mAP} \\
\centering \textbf{Cross\\validation\\setup} & \centering \textbf{Training\\dataset\\size} & \textbf{IoU / OKS} \\
 & & {50:95} & {75} & {50} & {50:95} \\
 & & \textbf{Scales} \\
 & & {all} & & & {s.} & {m.} & {l.} \\

\textit{k=10} & $\approx\mathrm{1985}$ & 0.54 & 0.60 & 0.82 & 0.40 & 0.56 & 0.62 \\
\textit{k=5} & $\approx\mathrm{1764}$ & 0.53 & 0.59 & 0.82 & 0.40 & 0.56 & 0.62 \\
\textit{k=2} & $\approx\mathrm{1103}$ & 0.52 & 0.58 & 0.80 & 0.39 & 0.54 & 0.60 \\

 & & {\textbf{Keypoint mAP}} \\

\textit{k=10} & $\approx\mathrm{1985}$ & 0.72 & 0.73 & 0.73 & 0.76 & 0.93 & 0.74 \\
\textit{k=5} & $\approx\mathrm{1764}$ & 0.72 & 0.73 & 0.73 & 0.76 & 0.92 & 0.74 \\
\textit{k=2} & $\approx\mathrm{1103}$ & 0.71 & 0.71 & 0.72 & 0.74 & 0.91 & 0.72 \\

\end{tblr}
\label{tab:k_fold_ablation}
\end{table}

To show how the different dataset sizes might affect fine-tuning performance, we utilize different cross-validation setups for YOLO11 Large training. The recipe is consistent with the first ablation; for $k=10$, even the same train test setup is used. For $k=5$ and $k=2$, we train the model on $k$ experiments with training and test datasets defined in Section~\ref{sec:eval}. The results of this experiment are shown in Table~\ref{tab:k_fold_ablation}. It can be seen that when only $\approx56\,\%$ of samples are used in the performance drop is at max $2.6\,\%$ $\mathrm{mAP}$. In addition, this shows that if the computational load of $k=10$ cross-validation experiments is too much, the $k=2$ cross-validation can be used instead, giving only slightly worse results.

\subsection{Effect of second annotation round on detection performance}

\begin{table}[b]
\centering
\caption{Difference in the bounding box and keypoint prediction performance when the YOLO11 model is trained on data from the first or second annotation round. The data amount increase is $\approx369\,\%$. Both models are evaluated as $k=10$ fold cross-validation.}
\scriptsize
\begin{tblr}{
    width=0.49\textwidth,
    colspec = {Q[c,f,1.7cm] | Q[c,f,1.5cm] | Q[c,m] | Q[c,m] | Q[c,m] | Q[c,m] | Q[c,m] | Q[c,m] },
    stretch = 0.3,
    rowsep = 1.1pt,
    vline{2,3} = {1}{text=\ },
    vline{2,3} = {8}{text=\ },
    vline{4,5} = {3}{text=\ },
    vline{7,8} = {5}{text=\ },
    hline{2,9,Z} = {1.2pt},
    hline{4,6} = {0.5pt},
    hline{7-8,10-Y} = {0.5pt,dashed},
    cell{1}{1} = {c = 2, r = 1}{c,m},
    cell{1}{3} = {c = 6, r = 1}{c,m},
    cell{2}{1} = {c = 1, r = 4}{r,m},
    cell{2}{2} = {c = 1, r = 4}{r,m},
    cell{2}{3} = {c = 6, r = 1}{c,m},
    cell{3}{6} = {c = 3, r = 1}{c,m},
    cell{4}{3} = {c = 6, r = 1}{c,b},
    cell{5}{3} = {c = 3, r = 1}{c,b},
    cell{8}{1} = {c = 2, r = 1}{c,f},
    cell{8}{3} = {c = 6, r = 1}{c,f},
}
\textbf{YOLO11 Large} &  & \textbf{Bounding box mAP} \\
\centering \textbf{Annotation\\round} & \centering \textbf{Training\\dataset\\size} & \textbf{IoU / OKS} \\
 & & {50:95} & {75} & {50} & {50:95} \\
 & & \textbf{Scales} \\
 & & {all} & & & {s.} & {m.} & {l.} \\
$2^\mathrm{nd}$ & $\approx\mathrm{1985}$ & 0.54 & 0.60 & 0.82 & 0.40 & 0.56 & 0.62 \\
$1^\mathrm{st}$ & $\approx\mathrm{538}$ & 0.47 & 0.52 & 0.78 & 0.36 & 0.50 & 0.55 \\
 & & {\textbf{Keypoint mAP}} \\
$2^\mathrm{nd}$ & $\approx\mathrm{1985}$ & 0.72 & 0.73 & 0.73 & 0.76 & 0.93 & 0.74 \\
$1^\mathrm{st}$ & $\approx\mathrm{538}$ & 0.67 & 0.68 & 0.68 & 0.71 & 0.88 & 0.69 
\end{tblr}
\label{tab:ds_version_ablation}
\end{table}
Similar to the previous experiment, since we have two different versions of the dataset labeled in two separate rounds, we can also compare how additional annotations improve detection performance. For model fine-tuning, we consider only $k=10$ cross-validation. The data is split such that each $i-th$ fold is guaranteed to have only data from a previous version of the same fold. We train the YOLO11~\cite{jocher_ultralytics_2024} model on both dataset versions. The results are available in Table~\ref {tab:ds_version_ablation}. As can be seen, the data increase of $\approx369\,\%$ improves performance by at least $22\,\%$ $\mathrm{mAP}$ and at most $47\,\%$ $\mathrm{mAP}$. This can provide a basis for any future extension of the Archival Faces dataset. There is still a significant gap of $46\,\%$ $\mathrm{mAP}$ for bounding boxes to be bridged. An additional increase in historical data might bring the detection closer to the performance of face detectors on contemporary data~\cite{li_asfd_2021} where $\mathrm{mAP}$ is upwards of $96\,\%$ (Wider Face medium difficulty).

\subsection{Performance of Detection Transformer Models}

\begin{table}[t]
    \centering
    \caption{Performance of detection transformers compared to convolutional neural network detectors. We fine-tuned pre-trained transformers according to $k=10$ cross-validation Archival Faces protocol, with the addition of the entire train portion of Wider Face~\cite{yang_wider_2016}. However, we mainly suspect the amount of data to cause poor test performance of these models.}
\scriptsize
\begin{tblr}{
    width=0.49\textwidth,
    colspec = {X[r, b] | Q[c,m] | Q[c,m] | Q[c,m] | Q[c,m] | Q[c,m] | Q[c,m] | Q[c,m] },
    stretch = 0.3,
    rowsep = 1.1pt,
    vline{2,3} = {1}{text=\ },
    vline{4,5} = {3}{text=\ },
    vline{7,8} = {5}{text=\ },
    hline{2,Z} = {1.2pt},
    hline{4,6} = {0.5pt},
    hline{7-Y} = {0.5pt,dashed},
    cell{1}{3} = {c = 6, r = 1}{c,m},
    cell{2}{1} = {c = 1, r = 4}{r,f},
    cell{2}{2} = {c = 1, r = 4}{r,f},
    cell{2}{3} = {c = 6, r = 1}{c,m},
    cell{3}{6} = {c = 3, r = 1}{c,m},
    cell{4}{3} = {c = 6, r = 1}{c,b},
    cell{5}{3} = {c = 3, r = 1}{c,b},
}

 & & \textbf{Bounding box mAP} \\
\textbf{Model} & \textbf{Variant} & \textbf{IoU} \\
 & & {50:95} & {75} & {50} & {50:95} \\
 & & \textbf{Scales} \\
 & & {all} & & & {s.} & {m.} & {l.} \\

\textit{YOLO11} & \textit{L} & 0.54 & 0.60 & 0.82 & 0.40 & 0.56 & 0.62 \\
\textit{YOLO11} & \textit{N} & 0.45 & 0.48 & 0.73 & 0.30 & 0.47 & 0.56 \\
\textit{YOLOV5-Face~\cite{qi_yolo5face_2023}} & \textit{M} & 0.24 & 0.19 & 0.52 & 0.20 & 0.22 & 0.33 \\
\textit{CoDETR} & \textit{-} & 0.08 & 0.03 & 0.23 & 0.03 & 0.10 & 0.17 \\
\textit{DETR} & \textit{-} & 0.05 & 0.02 & 0.17 & 0.03 & 0.07 & 0.13 \\
 
\end{tblr}
    \label{tab:detr}
\end{table}

In recent years, a paradigm shift in decoders has led to end-to-end transformer detectors, eliminating anchors and non-maxima suppression \cite{vedaldi_end--end_2020}. We include baselines for the performance of both the detection transformer (DETR)~\cite{vedaldi_end--end_2020} and the conditional DETR (CoDETR)~\cite{meng_conditional_2021}. Both models use the same training recipe, with images scaled to a maximum of $1333\times800$ pixels. In this case, data augmentation includes color jitter, random perspective, random rotation, Gaussian blur, and Gaussian noise. Similarly to YOLO models, we utilize COCO~\cite{lin_microsoft_2014} pre-trained weights. Similar to experiments with conv-nets, we append the entire training part of the Wider Face~\cite{yang_wider_2016} to our train sets. We evaluate models on $k=10$ cross-validation Archival Faces setup to have as much domain-specific training data as possible. Device batch size is set to 20 with eight gradient accumulation steps. Learning rate is $10^{-5}$, weight decay is $10^{-4}$, and we utilize AdamW~\cite{loshchilov_decoupled_2018} optimizer. Training is stopped after 15 epochs without any substantial improvement in validation loss after 10 epochs. The performance compared to fine-tuned conv-net detectors can be seen in Table~\ref{tab:detr}. As can be seen, the average precision is relatively poor on all metrics, not even reaching the best generic face detection models. We have validated our training pipeline, and the results seem consistent with fine-tuning the transformer on small datasets (below $10k$ samples)~\cite{avidan_towards_2022} without any significant architecture modifications.

\section{Conclusion}

The mission of this article was to shed light on the problem of face detection in photographs in archival documents. Existing face detectors work amazingly even for very challenging photographs taken and processed today but fail remarkably on photographs from digitized archives. We provide a publicly available dataset, Archival Faces, from this underrepresented domain with landmark annotations for alignment.

We trained face detectors based on YOLOv8~\cite{jocher_ultralytics_2023}, YOLO11~\cite{jocher_ultralytics_2024}, DETR~\cite{vedaldi_end--end_2020} and Conditional DETR~\cite{meng_conditional_2021} using the newly created dataset. We achieved good results: we show on our cross-validation protocol that after fine-tuning on Ar\-chi\-val Faces, YOLO11 can improve bounding box $\mathrm{mAP}$ for $\mathrm{IoU}$ $50:95$ from $22\,\%$  to $54\,\%$ compared to model trained only on Wider Face~\cite{yang_wider_2016}. We carried out experiments showing that even as little as $538$ Archival Faces images in training can give as much as $47\,\%$ $\mathrm{mAP}$ and that even $k=2$ cross-validation protocol can be used without significant performance hit.

Additional research may build on top of the Archival Faces by extending annotation counts or covering different unforeseen document-style domains. Alternatively, future work can utilize our dataset with a combination of conventional datasets (such as Wider Face) to style-transfer photos through procedural augmentations with Archival Faces as a style reference. The broader reach is towards the face recognition community, providing a substantial stepping stone for face recognition within historical archives. This may aid in creating new domain-specific face-recognition datasets and multimodal name-to-face links within documents.


\begin{credits}
\subsubsection{\ackname} This work has been supported by the Ministry of Culture Czech Republic in NAKI III project Orbis Pictus – book revival for cultural and creative sectors (DH23P03OVV033). This work was supported by the Ministry of Education, Youth and Sports of the Czech Republic through the e-INFRA CZ (ID:90254). We would like to especially thank our annotators, Martina Dvořáková (MZK)  and Petr Žabička (MZK), for their commitment and perseverance, which made this dataset possible. 

\subsubsection{\discintname}
The authors have no competing interests to declare relevant to this article's content.
\end{credits}

%
%
%
\bibliographystyle{splncs04}
\bibliography{references}
\end{document}